\documentclass[lettersize,journal]{IEEEtran}
\usepackage{amsmath,amsfonts}
\usepackage{algorithmic}
\usepackage{algorithm}
\usepackage{array}
\usepackage[caption=false,font=normalsize,labelfont=sf,textfont=sf]{subfig}
\usepackage{textcomp}
\usepackage{stfloats}
\usepackage{url}
\usepackage{verbatim}
\usepackage{graphicx}
\usepackage{cite}
\usepackage{multirow}
\usepackage{booktabs}
\usepackage{xcolor}
\usepackage[misc]{ifsym}

\begin{document}

\title{LORE++: Logical Location Regression Network for Table Structure Recognition with Pre-training}

\author{Rujiao Long$\,^{\ast}$, Hangdi Xing$\,^{\ast}$, Zhibo Yang, Qi Zheng, Zhi Yu, Cong Yao \textsuperscript{\Letter}, Fei Huang
%Rujiao Long, Hangdi Xing,  Zhi Yu, IEEE Publication Technology,~\IEEEmembership{Staff,~IEEE,}
        % <-this % stops a space
%\thanks{This paper was produced by the Alibaba Group. They are in Hangzhou, China.}% <-this % stops a space
%\thanks{Rujiao Long and Hangdi Xing contribute equally. }
\thanks{Rujiao Long, Zhibo Yang, Qi Zheng, Cong Yao and Fei Huang are with the Alibaba Group, Hangzhou, 310030, China (e-mail: rujiao.lrj@gmail.com, yangzhibo450@gmail.com, yongqi.zq@taobao.com, yaocong2010@gmail.com, feirhuang@gmail.com)}
%\thanks{Rujiao Long, Zhibo Yang, Fei Huang, Cong Yao and Qi Zheng are with the Alibaba Group, Hangzhou, China (e-mail: \{rujiao.lrj,zhibo.yzb,f.huang\}@alibaba-inc.com, yaocong2010@gmail.com, yongqi.zq@taobao.com)}
\thanks{Hangdi Xing and Zhi Yu are with the Zhejiang University, Hangzhou, 310027, China (e-mail: xinghd@zju.edu.cn, yuzhirenzhe@zju.edu.cn)}
\thanks{$\,^{\ast}$ Equal contribution. \Letter~Corresponding author: Cong Yao.}
}

% The paper h-maileaders
% \markboth{IEEE TRANSACTIONS ON IMAGE PROCESSING}%

% % column for its text to clear the IEEEpubid mark.

\maketitle

\begin{abstract}
Table structure recognition (TSR) aims at extracting tables in images into machine-understandable formats. Recent methods solve this problem by predicting the adjacency relations of detected cell boxes or learning to directly generate the corresponding markup sequences from the table images. However, existing approaches either count on additional heuristic rules to recover the table structures, or face challenges in capturing long-range dependencies within tables, resulting in increased complexity. In this paper, we propose an alternative paradigm. We model TSR as a logical location regression problem and propose a new TSR framework called LORE, standing for LOgical location REgression network, which for the first time regresses logical location as well as spatial location of table cells in a unified network. Our proposed LORE is conceptually simpler, easier to train, and more accurate than other paradigms of TSR. Moreover, inspired by the persuasive success of pre-trained models on a number of computer vision and natural language processing tasks, we propose two pre-training tasks to enrich the spatial and logical representations at the feature level of LORE, resulting in an upgraded version called LORE++. The incorporation of pre-training in LORE++ has proven to enjoy significant advantages, leading to a substantial enhancement in terms of accuracy, generalization, and few-shot capability compared to its predecessor. Experiments on standard benchmarks against methods of previous paradigms demonstrate the superiority of LORE++, which highlights the potential and promising prospect of the logical location regression paradigm for TSR.
\end{abstract}


\section{Introduction}
\IEEEPARstart{D}{ata} in tabular format is prevalent in various sorts of documents for summarizing and presenting information. As the world is going digital, the need for parsing the tables trapped in unstructured data (e.g., images and PDF files) is growing rapidly. Although straightforward for humans, it is challenging for automated systems due to the wide diversity of layouts and styles of tables. Table Structure Recognition (TSR) refers to transforming tables in images to machine-understandable formats, usually in logical coordinates or markup sequences. The extracted table structures are crucial for various applications, such as information retrieval, table-to-text generation, and question answering. 
%The extracted table structures are crucial for information retrieval, table-to-text generation, and question answering systems, etc. 

\begin{figure}[!t]
\centering
\captionsetup[subfloat]{font=scriptsize}
\subfloat[Adjacency relationship representations]{\includegraphics[width=0.96\linewidth]{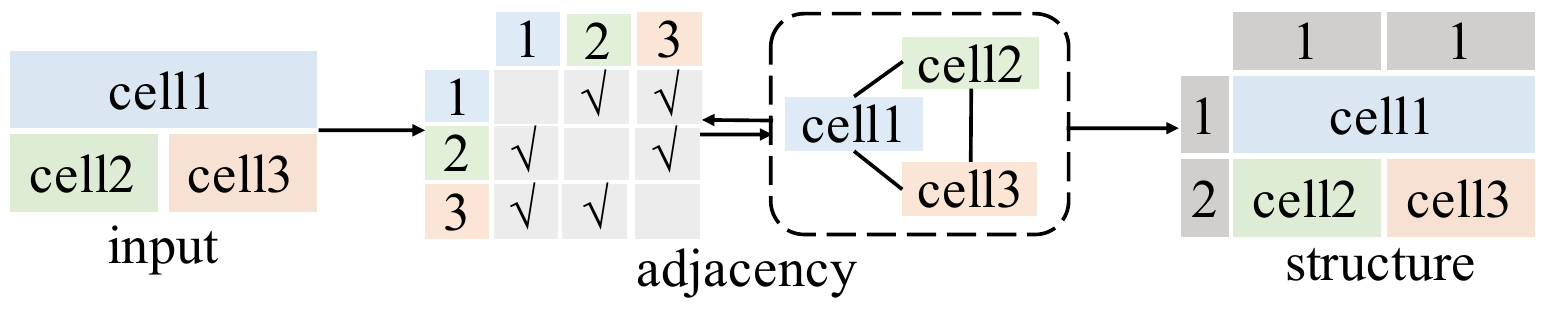}%
\label{fig1_first_case}}
\hfil
\subfloat[Markup sequence representations]{\includegraphics[width=0.96\linewidth]{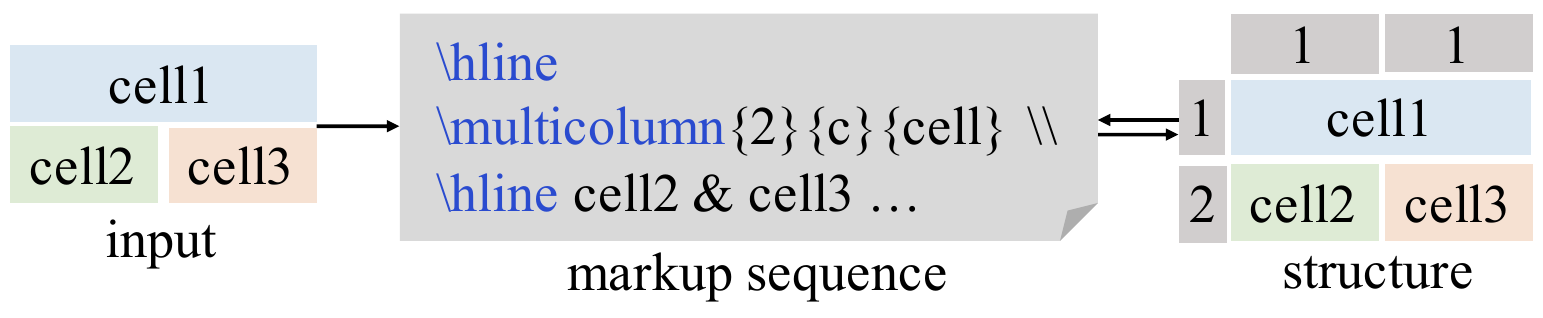}%
\label{fig1_second_case}}
\hfil
\subfloat[Logical location representations]{\includegraphics[width=0.96\linewidth]{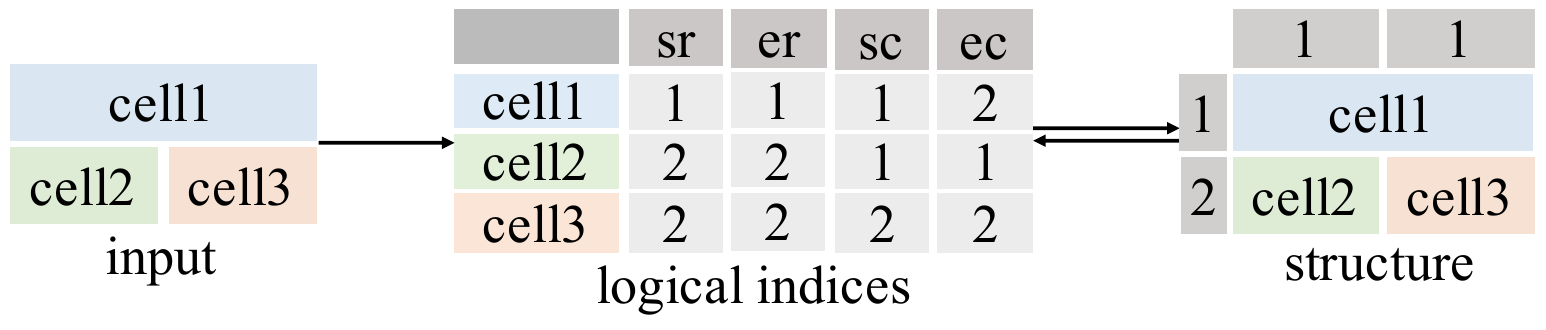}%
\label{fig1_third_case}}
\caption{TSR paradigms using different table-structure representations. Here, $sr$, $er$, $sc$, $ec$ refer to the starting-row, ending-row, starting-column, and ending-column respectively.}
\label{fig_sim}
\end{figure}

Pioneer methods \cite{raja2022visual, long2021parsing, qiao2021lgpma, zheng2021global, prasad2020cascadetabnet, paliwal2019tablenet} elaborately design the detectors to accurately obtain the spatial locations, i.e., bounding boxes of table cells, and recover the table structure by heuristic rules based on visual clues including lines, aligned cell boundaries and text regions. With the development of deep learning, TSR methods have recently advanced substantially by automatically predicting the structure of the table. Most deep learning-based TSR methods can be categorized into the following paradigms. The first type of models \cite{chi2019complicated, raja2020table, liu2021neural} aim at exploring the adjacency relationships between pairs of detected cells to generate intermediate results. They rely on tedious post-processing or graph optimization algorithms to reconstruct the table as logical coordinates, as depicted in Figure \ref{fig1_first_case}, which would struggle with complex table structures. Another paradigm formulates TSR as a markup language sequence generation problem \cite{zhong2020image, desai2021tablex}, as shown in Figure \ref{fig1_second_case}. Although it simplifies the TSR pipelines, the models are supposed to redundantly learn a markup grammar from noisy sequence labels, which results in a much larger amount of training data. Besides, these models are time-consuming due to the sequential decoding process.

\begin{figure}[t]
  \centering
\includegraphics[width=0.97\columnwidth]{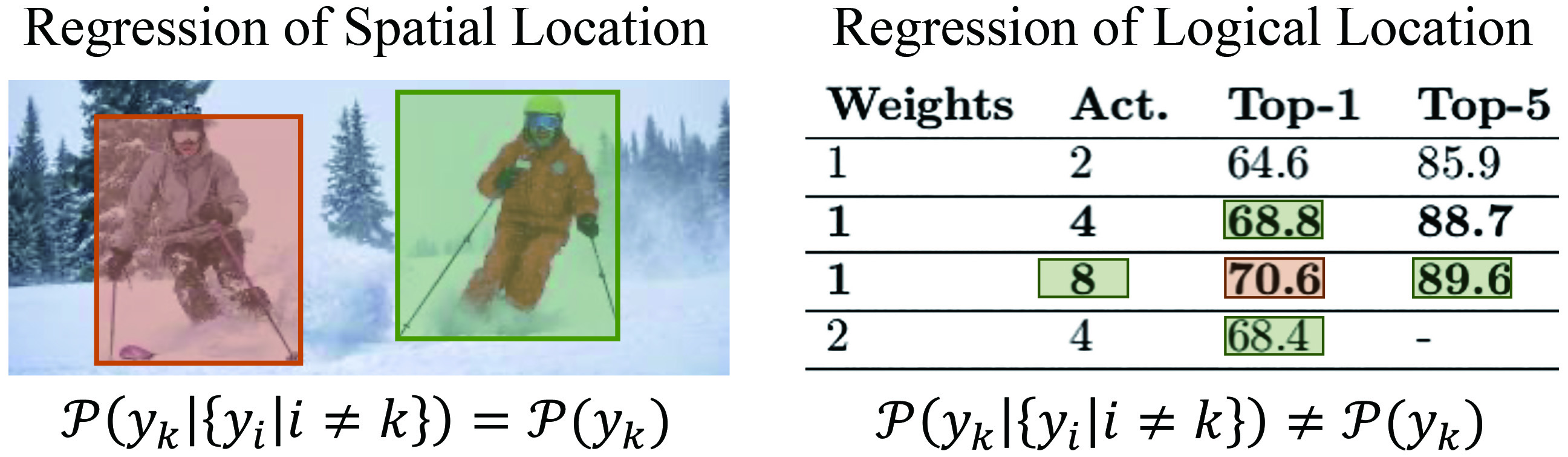}    

\caption{A comparison between the usual regression (left) and the logical location regression (right). The typical regression hypothesis is that different targets are independently distributed. However, dependencies exist between logical indices, e.g., the logical location of the cell `70.6' is constrained by those of the four surrounding cells.}
\label{fig:log}
\end{figure}

In fact, logical coordinates are well-defined machine-understandable representations of table structures, which are complete to reconstruct tables and can be converted into adjacency matrices and markup sequences by simple and clear transformations. Recently, there has been a focus on exploring the logical locations of table cells \cite{xue2021tgrnet} as depicted in Figure \ref{fig1_third_case}. However, the method predicts logical locations by ordinal classification, which is apt to suffer from the long-tailed distribution of row (column) numbers. More importantly, this method does not account for the natural dependencies between logical locations. For example, the design of a table itself is from top to bottom, left to right, causing the logical location of cells to be interdependent. This nature of logical locations is sketched in Figure \ref{fig:log}. Furthermore, the work lacks a comprehensive comparison among various TSR paradigms.

Aiming at breaking the limitations of existing methods, we propose \textbf{LO}gical Location \textbf{RE}gression Network (LORE for abbreviation), a conceptually simpler and more effective TSR framework. It first locates table cells on the input image and then predicts the logical locations along with the spatial locations of cells. To better model the dependencies and constraints between logical locations, a cascade regression framework is adopted, combined with the inter-cell and intra-cell supervisions. The inference of LORE is a parallel network forward-pass, without any efforts in complicated post-processings or sequential decoding strategies.

We evaluate LORE on a wide range of benchmarks against TSR methods of different paradigms. Experiments show that LORE is highly competitive and outperforms previous state-of-the-art methods. Specifically, LORE surpasses other logical location prediction methods by a large margin. Moreover, the adjacency relations and markup sequences derived from the predictions of LORE are of higher quality, which demonstrates that LORE covers the capacity of the models trained under other TSR paradigms.

% Back to the ability of understanding table of human, we can easily grasp the structure of tables regardless of their layouts and styles, because we are good at 
Despite the advancements in TSR through the logical location regression paradigm, it does not leverage the abundance of tables from larger datasets \cite{zhong2020image, smock2022pubtables}. Inspired by the persuasive success of pre-trained models on both computer vision and natural language processing tasks \cite{devlin2018bert, He2021MaskedAA}, we extend the LORE model to a pre-trained version, i.e. LORE++. Humans can grasp the structure of tables effortlessly regardless of the various layouts and styles since the awareness of basic vision clues and the clear notion of logical grids in mind. So we dedicated ourselves to enlightening the model to learn the logical row and column grids. In fact, the tasks of spatial and logical location prediction are closely intertwined. Enhancing the accuracy of the spatial location prediction will result in improved precision in the logical location prediction. Therefore, we employ the Masked Autoencoder (MAE) task to enhance the model's understanding of tabular images, thereby improving the outcomes of spatial location prediction. Additionally, we propose a novel pre-training task called Logical Distance Prediction (LDP) to comprehend the logical relationships between text segments in the images, thereby boosting the model's capability in logical location prediction. Besides, we utilize a masking strategy that enables the joint training of two pre-training tasks in a single forward-pass. We have curated a pre-training dataset of 1.5 million samples by integrating academic datasets and generated labels for the LDP task by extracting OCR results using an open-source OCR engine. Consequently, LORE++ improves the vanilla LORE in terms of accuracy and data efficiency.
%Specifically, we propose to jointly pretrain the vision encoder and the logical location prediction head with two tasks. First, the vision module is pretrained by the Masked Autoencoder (MAE) task similar to \cite{zhong2020image}. Second, the overall LORE architecture is pretrained by the Logical Distance Prediction task, which request the model to predict the row and column difference of cells. Consequently, LORE++ improves the vanilla LORE in terms of accuracy and data efficiency.

Our main contributions can be summarized as follows:
\begin{itemize}
    \item We propose to model TSR as a logical location regression problem and design LORE, a new TSR framework that captures dependencies and constraints between logical locations of cells, and predicts the logical locations along with the spatial locations.
    
    \item We empirically demonstrate that the logical location regression paradigm is highly effective and covers the abilities of previous TSR paradigms, such as predicting adjacency relations and generating markup sequences.
    
    % \item LORE outperforms previous methods on standard benchmarks. The code of LORE will be made publicly available, providing a hands-off way to apply an effective TSR model.
    \item We extend LORE to LORE++ by introducing two pre-training tasks specially designed for TSR. LORE++ can extract enhanced representations via the two pre-training tasks, which lead to improved accuracy and better few-shot capability, compared with its predecessor. 
\end{itemize}

The following of this paper is organized as follows. Section \uppercase\expandafter{\romannumeral 2} describes the relevant works to our paper and preliminaries of TSR. We then detail the architecture of the vanilla LORE model as well as the pre-train framework design for LORE++ in Section \uppercase\expandafter{\romannumeral 3}, \uppercase\expandafter{\romannumeral 4}, and Section \uppercase\expandafter{\romannumeral 5}. Next, Section \uppercase\expandafter{\romannumeral 6} presents the extensive experimental results and analyses. Finally, we conclude our paper in Section \uppercase\expandafter{\romannumeral 7} and discuss the potential applications and future research directions.

The LORE was originally proposed in our previous conference paper \cite{xing2023lore}. This article extends that work with the following improvements and modifications: (1) In order to further improve the performance and generalization ability of LORE, we extend it into a pre-trained version, termed LORE++. We design a pre-training framework to jointly train the logical location regression network on both spatial and logical tasks (see Section \uppercase\expandafter{\romannumeral 5}). (2) We conduct comprehensive experiments and demonstrate that LORE++ can substantially boost both the cell detection and structure recognition results. More importantly, it shows superiority in terms of data efficiency and generalization ability.
(3) We also present more ablation studies, comparisons and analyses. Besides, to better facilitate real-world applications, we devise and realize the transformations from logical locations to relation adjacent matrices and markup sequences (see Section \uppercase\expandafter{\romannumeral 6}).

\section{Related Work}
\subsection{TSR based on accurate cell segmentation}
Early works \cite{schreiber2017deepdesrt, siddiqui2019deeptabstr} introduce segmentation or detection frameworks to locate and extract splitting lines of table rows and columns. Subsequently, they reconstruct the table structure by empirically grouping the cell boxes with pre-defined rules. These models would suffer from tables with spanning cells or distortions. The latest baselines \cite{long2021parsing, smock2022pubtables, zhang2022split} tackle this problem with well-designed detectors or attention-based merging modules to obtain more accurate cell boundaries and merging results.  However, they either are tailored for a certain type of datasets or require customized processing to recover table structures, and thus can hardly be generalized. So there arise models focusing on directly predicting the table structures with neural networks.

\subsection{TSR based on directly structure prediction}

\cite{chi2019complicated} proposes to model table cells as text segmentation regions and exploit the relationships between cell pairs. Precisely, it applies graph neural networks \cite{kipf2016semi} to classify pairs of detected cells into horizontal, vertical, and unrelated relations. Following this work, there are models devoted to improving the relationship classification by using elaborated neural networks and adding multi-modal features \cite{qasim2019rethinking, raja2020table, raja2022visual, liu2021show, liu2021neural}. This framework bypasses the extraction of precise boundaries of cells, but the nearest neighbor graphs in these models encode biased prior to the model. Moreover, there is still a gap between the set of relation triplets and the global table structure. Complex graph optimization algorithms or pre-defined post-processings are needed to recover the tables.

\cite{li-etal-2020-tablebank, zhong2020image, ye2021pingan} make the pioneering attempts to solve the TSR problem in an end-to-end way. They employ sequence decoders to generate tags of markup language that represent table structures. However, the models are supposed to learn the markup grammar with noisy labels, resulting in the methods being difficult to train and requiring tens of times more training samples than methods of other paradigms. Besides, these models are time-consuming owing to the sequential decoding process.

\cite{xue2021tgrnet} propose to perform ordinal classification of logical indices on each detected cell for TSR, which is close to our approach. The model utilizes graph neural networks to classify detected cells into the corresponding logical locations, while it ignores the dependencies and constraints among logical locations of cells. Besides, the model is only evaluated on a few datasets and not against the strong TSR baselines. 

% \cite{xue2021tgrnet} propose to perform ordinal classification of logical indices on each detected cell for TSR, which is close to our approach. The model utilizes graph neural networks to classify detected cells into the  corresponding logical locations, while it ignores the dependencies and constraints among logical locations of cells. Besides, the model is only evaluated on a few datasets and not against the strong TSR baselines. 

\subsection{Pre-training models}
The pre-trained model shows remarkable performance on numerous CV tasks, e.g. classification\cite{He2021MaskedAA}, segmentation\cite{xu2022groupvit, xu2023open}, detection\cite{dai2021up, bar2022detreg}, and information extraction\cite{luo2023geolayoutlm, yang2023modeling}, which proved the pre-trained representations generalize well to various downstream data. Unfortunately, there is currently no pre-training work in TSR with images as input only.

MAE\cite{He2021MaskedAA} masks random patches of the input image and reconstructs the missing pixels. Successfully reconstructing the object in the image indicates that the model understands what the object is and what it looks like. So, we utilize MAE as one of our pre-training tasks to absorb various layout and structure information from massive data.

Besides, ESP\cite{yang2023modeling} constructs key-value linking task in pre-training to model entity linking task in fine-tuning. Due to the consistency between pre-training and fine-tuning, ESP improves the SOTA accuracy of linking tasks from 81.25\% to 92.31\% on XFUND\cite{wang2021towards} dataset. Inspired by ESP, we use text segments instead of table cells to model logical relationship, for which the text segments can be easily obtained by the OCR engine. 

%Our LORE++ borrows the idea of pre-trained models to enable LORE to generalize on the diversified design and layout of tables. Masked image modeling, represented by MAE, is one of the latest self-supervised learning strategies. As a pre-training framework, masked autoencoders have shown a broad impact on visual recognition. However, original masked autoencoders are not directly applicable to the Logical  uses a few convolutional blocks as input tokenizers. To the best of our knowledge, there are no pretrained models that show 

%\subsection{Comparison to the Conference Version}
%In preliminary conference work, i.e. LORE, was presented in \cite{xing2023lore}. This journal paper extends the previous study with the following improvements and modifications:

%(1) In order to further improve the performance and generalization ability of LORE, we extend it into a pre-trained version, termed as LORE++. We design a pre-training framework to jointly train the logical location regression network on both spatial and logical tasks. 

%(2) We conduct comprehensive experiments and demonstrate that the LORE++ can substantially boost both the cell detection and structure recognition results. More importantly, it shows superiority in terms of data efficiency and generalization ability.

%(3) We also present more ablation studies and experiment
%analysis, inference speed, etc.. Besides, to better facilitate application, we specify the transformation for logical location to relation adjacent matrix and markup sequence.

\section{Preliminaries}
\subsection{Problem Definition}
In this paper, we consider the TSR problem as the spatial and logical location regression task. Specifically, for an input image of the table, similar to a detector, a set of table cells $\{O_1,O_2,...,O_N\}$ are predicted as their logical locations $\{l_1, l_2, ..., l_N\}$, along with the spatial locations $\{B_1, B_2, ..., B_N\}$, where $l_i = (r_s^{(i)}, r_e^{(i)}, c_s^{(i)}, c_e^{(i)})$ standing for the starting-row, ending-row, starting-column and ending-column, $B_i = \{(x_k^{(i)}, y_k^{(i)})\}_{k=1,2,3,4}$ standing for the four corner points of the $i$-th cell and $N$ is the number of cells in the image.

With the predicted table cells represented by their spatial and logical locations, the table in the image can be converted into machine-understandable formats, such as relational databases. Besides, the adjacency matrices and the markup sequences of tables can be directly derived from their logical coordinates with well-defined transformations rather than heuristic rules (See supplementary section 1).

\subsection{Transformation on Logical Coordinates }
The transformation from the table representation in the logical location of cells to cell adjacency and markup sequence representation is well-defined for general settings, without any approximation algorithm or heuristic rules.

\begin{algorithm}[tb]
\caption{From logical location to markup sequence}
\label{alg:algorithm}
\textbf{Input}: \{$C_1, ..., C_K$\}, markup = ' '\\
\textbf{Output}: markup 
\begin{algorithmic}[1] %[1] enables line numbers
\STATE Let $i=0$.
\FOR{$i=0$ to $K-1$ }
\STATE markup += '\textless tr\textgreater'
\FOR{$j=0$ to $|C_k|-1$ }
\STATE $a_{ik} = (C_k)_i$
\STATE $rsp = 1 + r_e^{(a_{ik})} - r_s^{(a_{ik})}$
\STATE $csp = 1 + c_e^{(a_{ik})} - c_s^{(a_{ik})}$
\STATE markup +=  \\
\textless td rowspan = $rsp$ colspan = $csp$ \textgreater \textless/td\textgreater'
\ENDFOR
\STATE markup += \textless /tr\textgreater
\ENDFOR

\STATE \textbf{return} markup
\end{algorithmic}
\end{algorithm}

\subsubsection{Adjacency from Logical Location}
The definition of adjacency of cells is based on the logical location \cite{gobel2012methodology}. Given a set of cells represented by logical locations as $C = \{a | a=(r_s, r_e, c_s, c_e) \}$, where $r_s, r_e, c_s, c_e$ denote the starting-row, ending-row, starting-column, the adjacency between two cells $a,b \in C$ is a binary relationship $R$ as:

\begin{equation}
    aRb := (p \land q )\vee(r \land s),
\end{equation}
where proposition $p,q,r,s$ are defined as:

\begin{equation}
\begin{aligned}
    p: (r_e^{(b)}<= r_s^{(a)} <= r_e^{(b)}) \vee \\
      (r_e^{(a)}<= r_s^{(b)} <= r_e^{(a)}), %same row
\end{aligned}
\end{equation}

\begin{equation}
    q: (c_s^{a} - c_e^{(b)} = 1) \vee (c_s^{b} - c_e^{(a)} = 1), %next column
\end{equation}
where $p \land q$ denotes $a$ and $b$ are locating (spanning) in the same row and a/b is exactly in the next column of b/a.
\begin{equation}
\begin{aligned}
    r: (c_e^{(b)}<= c_s^{(a)} <= c_e^{(b)}) \vee \\
    (c_e^{(a)}<= c_s^{(b)} <= c_e^{(a)}), %same column
\end{aligned}
\end{equation}

\begin{equation}
    s: (r_s^{a} - r_e^{(b)} = 1) \vee (r_s^{b} - r_e^{(a)} = 1). %next row
\end{equation}
And $r\land s$ is defined similar as $p \land q$. By this definition, the adjacency of cells can be straightforwardly computed from the logical location of cells.

\subsubsection{Markup from Logical Location}
Given a table represented as logical coordinates, we first define the table rows as finite ordered sets, following the notations before as:

\begin{equation}
    C_k = \{a_i | r_s^{(a_i)}=k, c_s^{(a_1)}<c_s^{(a_1)}<...<c_s^{(a_N)} \},
\end{equation}
where $C_k$ denotes the set of cells in the $k$-th row of the table and $N=|C_k|$. These sets exist and are subject to :
\begin{equation}
    C = C_1 \cup C_2 \cup ... \cup C_K ,\\
\end{equation}
and:
\begin{equation}
     C_i \cap C_j = \emptyset, i,j \in \{1,2,...,K\} ,%same column
\end{equation}

Where $K$ is the number of rows. The transformation of logical coordinates into markup sequence is then defined as in the algorithm \ref{alg:algorithm}.

\section{Model Architecture}
\begin{figure*}[t]
  \centering
\includegraphics[width=1.98\columnwidth]{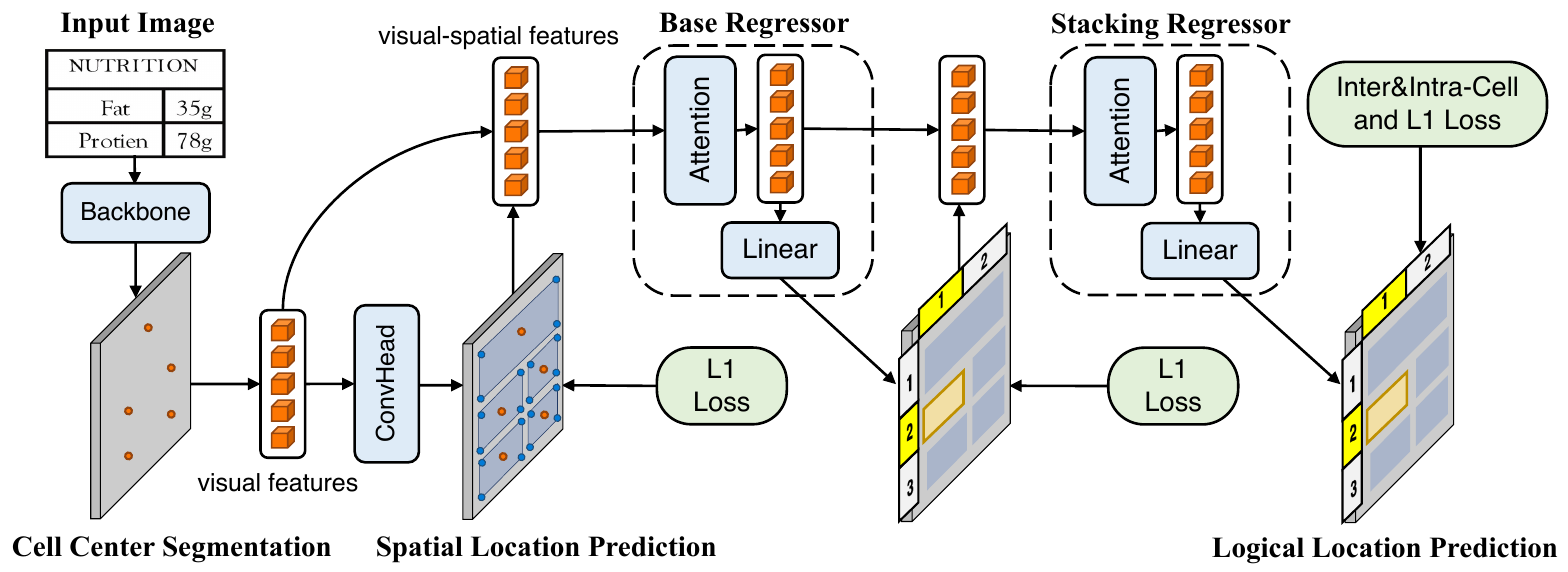}    

\caption{An illustration of LORE. It first locates table cells in the input image by key point segmentation. Then the logical locations are predicted along with the spatial locations. The cascading regressors and the inter-cell and intra-cell supervisions are employed to better model the dependencies and constraints between logical locations. }
\label{fig:lore}
\end{figure*}

This section elaborates on our proposed LORE, a TSR framework regressing the spatial and logical locations of cells. As illustrated in Figure \ref{fig:lore}, LORE employs a vision backbone to extract visual features of table cells from the input image. Then the spatial and logical locations of cells are predicted by two regression heads. We specially leverage the cascading regressors and employ inter-cell and intra-cell supervisions to model the dependencies and constraints between logical locations. The following subsections specify these crucial components. 

% The pre-training framework is sketched in Figure . We leverage the MAE task to enhance the vision backbone and the cell relation prediction task to enlighten the model to encode the notion of logical row and column. The following subsections specify these crucial components and pre-training task respectively. 

\subsection{Table Cell Features Preparation}
In order to streamline the joint prediction of spatial and logical locations, we employ a key point segmentation network \cite{zhou2019objects,long2021parsing} as the feature extractor and model each table cell in the image as its center point. Besides, it is compatible with both wired and wireless tables and is easier to implement on inconsistent annotations of different datasets, i.e., aligned boxes in WTW and TG24, or text region boxes in SciTSR and PubTabNet. 

For an input image of width $W$ and height $H$, the network produces a feature map $f \in \mathbb{R}^{\frac{W}{R} \times \frac{H}{R}\times d} $ and a cell center heatmap $\widehat{Y} \in [0,1]^{\frac{W}{R} \times \frac{H}{R}} $, where $R$, $d$ are the output stride and hidden size; $ \widehat{Y}_{x,y} = 1 $ corresponds to a detected cell center, while $ \widehat{Y}_{x,y} = 0 $ refers to the background. 

In the subsequent modules, the CNN features $\{f^{(1)}, f^{(2)},..., f^{(N)}\}$ at detected cell centers $\{\hat{p}^{(1)}, \hat{p}^{(2)}, ..., \hat{p}^{(N)}\}$ are considered as the representations of table cells.

\subsection{Spatial Location Regression} 
 
We choose to predict the four corner points rather than the rectangle bounding box to better deal with the inclines and distortions of tables in the wild. For spatial locations, the features of the backbone $f$ are passed through a $3 \times 3$ convolution, ReLU and another $1 \times 1$ convolution to get the prediction $\{\hat{B}^{(1)}, \hat{B}^{(2)}, ..., \hat{B}^{(N)}\}$ on centers $\{\hat{p}^{(1)}, \hat{p}^{(2)}, ..., \hat{p}^{(N)}\}$, where $\hat{B}^{(i)} = \{(\hat{x}_k^{(i)}, \hat{y}_k^{(i)})\}_{k=1,2,3,4}$.

\subsection{Logical Location Regression}
As dense dependencies and constraints exist between the logical locations of table cells, it is rather challenging to learn the logical coordinates from the visual features of cell centers alone. The cascading regressors with inter-cell and intra-cell supervisions are leveraged to explicitly model the logical relations between cells.

\subsubsection{Base Regressor}
To better model the logical relations from images, the visual features are first combined with the spatial information. Specifically, the features of the predicted corner points of the cells are computed as the sum of their visual features and 2-dimensional position embeddings:

\begin{equation}
    \widetilde{f}_{(\hat{x}_k^{(i)},\hat{y}_k^{(i)},:)} = 
f_{(\hat{x}_k^{(i)},\hat{y}_k^{(i)},:)} + PE(\hat{x}_k^{(i)},\hat{y}_k^{(i)}),
\end{equation}
where $PE$ refers to the 2-dimensional position embedding function \cite{xu2020layoutlm, xu2020layoutlmv2}. Then the features of the four corner points are added to the center features $f^{(i)}$ to enhance the representation of each predicted cell center $\hat{p}^{(i)}$ as:
\begin{equation}
    h^{(i)} = f^{(i)} + \sum_{k=1}^4 w_k \widetilde{f}_{(\hat{x}_k^{(i)},\hat{y}_k^{(i)},:)},
\end{equation}
where $[w_1, w_2, w_3, w_4]$ are learnable parameters.

Then the message-passing and aggregating networks are adopted to incorporate the interaction between the visual-spatial features of cells:
% \cite{qasim2019rethinking, raja2020table, xue2021tgrnet}
\begin{equation}
    \{ \widetilde{h}^{(i)}\}_{i = 1,2,...,N} = {\rm \textbf{Self\-Attention}}(\{ h^{(i)}\}_{i = 1,2,...,N}).
\end{equation} 

We use the self-attention mechanism \cite{vaswani2017attention} in LORE to avoid making additional assumptions about the distribution of table structure, rather than graph neural networks employed by previous methods \cite{qasim2019rethinking, xue2021tgrnet}, which will be further discussed in experiments.

The prediction of the base regressor is then computed by a linear layer with the ReLU activation from $\{ \widetilde{h}^{(i)}\}_{i = 1,2,...,N} $ as $\hat{l}^{(i)} = (\hat{r}_{s}^{(i)}, \hat{r}_{e}^{(i)}, \hat{c}_{s}^{(i)}, \hat{c}_{e}^{(i)})$.

\subsubsection{Stacking Regressor}
\label{sec:reg}
Although the base regressor encodes the relationships between visual-spatial features of cells, the logical locations of each cell are still predicted individually. To better capture the dependencies and constraints among logical locations, a stacking regressor is employed to look again at the prediction of the base regressor. Specifically, the enhanced features $\boldsymbol{\widetilde{h}}$ and the logical location prediction of the base regressor $\hat{\boldsymbol{l}}$ are fed into a stacking regressor. The stacking regressor can be expressed as :

\begin{equation}
 \widetilde{\boldsymbol{l}} = F_s(W_s\hat{\boldsymbol{l}}+ {\boldsymbol{\widetilde{h}}}).
\end{equation}

where $W_s \in \mathbb{R}^{4\times d}$ is a learnable parameter,  $ \hat{\boldsymbol{l}} = [\hat{l}^{(1)},...,\hat{l}^{(N)}]$, $ {\boldsymbol{\widetilde{h}}} = [\widetilde{h}^{(1)},...,\widetilde{h}^{(N)}]$ and $F_s$ denotes the stacking regression function, which has the same self-attention and linear structure as the base regression function but with independent parameters. The output of the stacking regressor is $ \widetilde{{\boldsymbol{l}}} = [\widetilde{l}^{(1)},...,\widetilde{l}^{(N)}]$, and $\widetilde{l}^{(i)} = (\widetilde{r}_{s}^{(i)}, \widetilde{r}_{e}^{(i)}, \widetilde{c}_{s}^{(i)}, \widetilde{c}_{e}^{(i)})$.

At the inference stage, the results are obtained by assigning the four components of $\widetilde{l}^{(i)}$ to the nearest integers. 

\subsubsection{Inter-cell and Intra-cell Supervisions}
In order to equip the logical location regressor with a better understanding of the dependencies and constraints between logical locations, we propose the inter-cell and intra-cell supervisions, which are summarized as: 1) The logical locations of different cells should be mutually exclusive (inter-cell). 2) The logical locations of one table cell should be consistent with its spans (intra-cell).

In practice, predictions of cells that are far apart rarely contradict each other, so we only sample adjacent pairs for inter-cell supervision. More formally, the scheme of inter-cell and intra-cell losses can be expressed as:

\begin{equation}
\begin{aligned}
    L_{inter} & = \sum_{(i,j) \in A_r}max(\widetilde{r}_e^{(j)} - \widetilde{r}_s^{(i)} + 1,  0) \\
    & + \sum_{(i,j) \in A_c}max(\widetilde{c}_e^{(j)} - \widetilde{c}_s^{(i)} + 1, 0),
\end{aligned}
\end{equation}
where $A_r$ ($A_c$) are sets of ordered horizontally (vertically) adjacent pairs, i.e., for a pair of cells $(i,j) \in A_r$ ($A_c$), cell $i$ is adjacent to cell $j$ in the same row (column) and on the right of (under) cell $j$, and $\widetilde{r}_s^{(i)}$,  $\widetilde{r}_e^{(j)}$, $\widetilde{c}_s^{(i)}$, $\widetilde{c}_e^{(j)}$ are predicted logical indices of cell $i$ and cell $j$. 
% where $A_r = \{(i,j)| \text{i,j are row-adjacent} and r_s^{(i)} - r_e^{(j)} = 1\}$ and $A_c = \{(i,j)|c_s^{(i)} - c_e^{(j)} = 1\}$ are the sets of ordered adjacent pairs. 

\begin{equation}
\begin{aligned}
    L_{intra} & = \sum_{i \in M_r}|\widetilde{r}_s^{(i)} - \widetilde{r}_e^{(i)}  - r_s^{(i)} + r_e^{(i)}| \\ 
    & + \sum_{i \in M_c}|\widetilde{c}_s^{(i)} - \widetilde{c}_e^{(i)} - c_s^{(i)} + c_e^{(i)} |,
\end{aligned}
\end{equation}
where $M_r = \{i|r_e^{(i)} - r_s^{(i)} \neq 0\}$ and $M_c = \{i|c_e^{(i)} - c_s^{(i)} \neq 0\}$ are sets of multi-row and multi-column cells. 

Then the inter-cell and intra-cell losses (I2C) are as:
$$L_{I2C} = L_{inter} + L_{intra}.$$

The supervisions are conducted on the output $\widetilde{{\boldsymbol{l}}}$ and no extra forward-passing is required.

\begin{figure*}[t]
  \centering
\includegraphics[width=1.98\columnwidth]{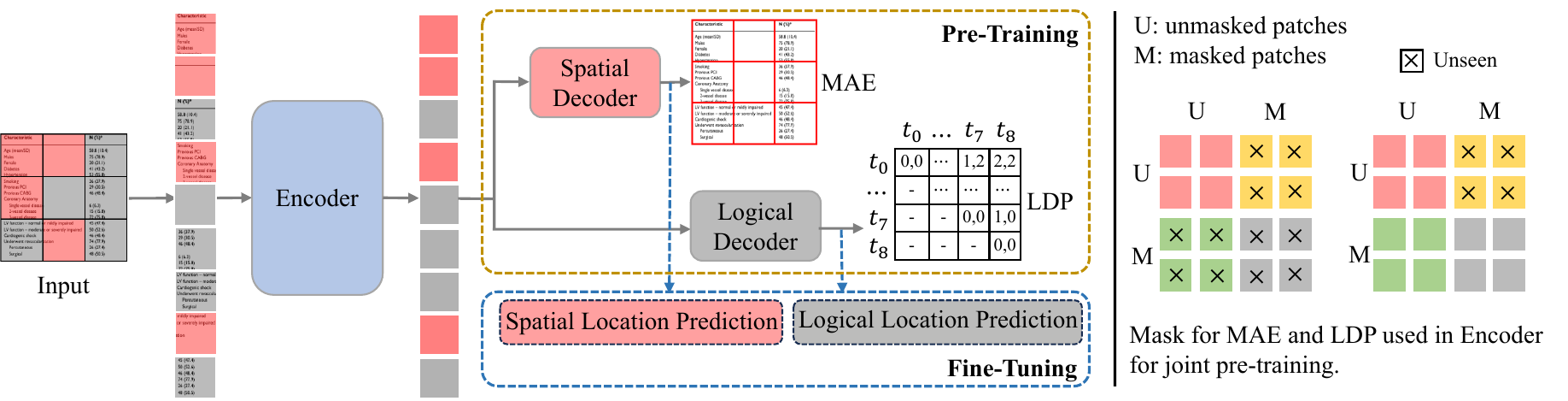}    

\caption{(Left) An illustration of the pre-training and fine-tuning framework of LORE++. The model is jointly pre-trained by the MAE task and the Logical Distance Prediction (LDP) task, which respectively corresponds to the spatial and logical location prediction task in the fine-tuning stage. (Right) The comparison of attention mask for MAE and LDP used in the encoder, which facilitates the model updating both the unmasked and masked patches in a single forward-pass. The embedding of masked patches will be replaced by the mask token when inputted into the spatial decoder.}
%Using two types of masks, unmasked patches are prevented from seeing masked patches, thereby avoiding data leakage that would render MAE training ineffective. At the same time, masked patches are still able to be encoded properly, ensuring the normal training of the LDP task.
% The comparison of vanilla attention mask and the unidirectional attention mask, which facilitate the model updating both the unmasked and masked patches in a single forward-pass.
\label{fig:lorev2}
\end{figure*}

\subsection{Objectives}
The losses of cell center segmentation $L_{center}$ and spatial location regression $L_{spa}$ are computed following typical key point-based detection methods \cite{zhou2019objects,long2021parsing}. 

The loss of logical locations is computed for both the base regressor and the stacking regressor:
\begin{equation}
     L_{log} =  \frac{1}{N} \sum_{i=1}^N (||\hat{l}^{(i)} - l_i||_1 + ||\widetilde{l}^{(i)} - l_i||_1).
\end{equation}

The total loss of joint training is then computed by adding the losses of cell center segmentation, spatial and logical location regression along with the I2C supervisions:
\begin{equation}
     L_{LORE} = L_{center} + L_{spa} + L_{log} + L_{I2C}.
\end{equation}

\section{Pre-training Framework}
We present a pre-training framework with objectives that include both visual and logical structure of tables, which enhances the model's ability to comprehend and reason about table structure in a holistic manner. In order to conduct joint training in a single forward pass and avoid information leaks, we propose unidirectional self-attention. Figure \ref{fig:lorev2} shows the pre-training framework and self-attention masking strategy. 

\subsection{Model Architecture}
As illustrated in Figure \ref{fig:lorev2}, We follow the framework of LORE with corresponding modifications to facilitate the joint pretraining of spatial and logical tasks. The CNN-based backbone of LORE is replaced with the ViT encoder and the MAE ViT-based decoder is added, catering to the MAE-like pre-training. The logical decoder shares a similar structure with the base regressor and the stacking regressor, i.e. layers of self-attention mechanism, but with the linear map on paired features for the logical distance prediction task. 

The model takes the patchified images and the corresponding masks as input. Aiming at preventing information leakage from the masked patches to the unmasked ones, we leverage a unidirectional self-attention, where the unmasked tokens can only attend to each other but not the masked ones, while the masked patches can attend to all patches.  The masked patches are replaced with the mask token before being inputted into the spatial decoder, while the entire encoded feature maps are forwarded to the logical decoder for the Logical Distance Prediction task.

\subsection{Masked Autoencoding Task}
Inspired by ViT-MAE, we utilize the MAE task to guide the model to learn general visual clues of tables such as text region, ruling lines, etc. We mask out 50\% patch tokens of the image randomly, which is different from the fashion as was suggested in ViT-MAE. Because table images are highly semantic and information-dense, excessively masking out patches could impede the reconstruction task. To introduce order information for the MAE task, we employ 1-D fixed sinusoidal position embeddings. The image encoder and decoder are trained using a normalized mean squared error (MSE) pixel reconstruction loss, which quantifies the disparity between the normalized target image patches and the reconstructed patches. This loss is specifically computed for the masked patches.

\begin{figure}[t]
  \centering
\includegraphics[width=0.93\columnwidth]{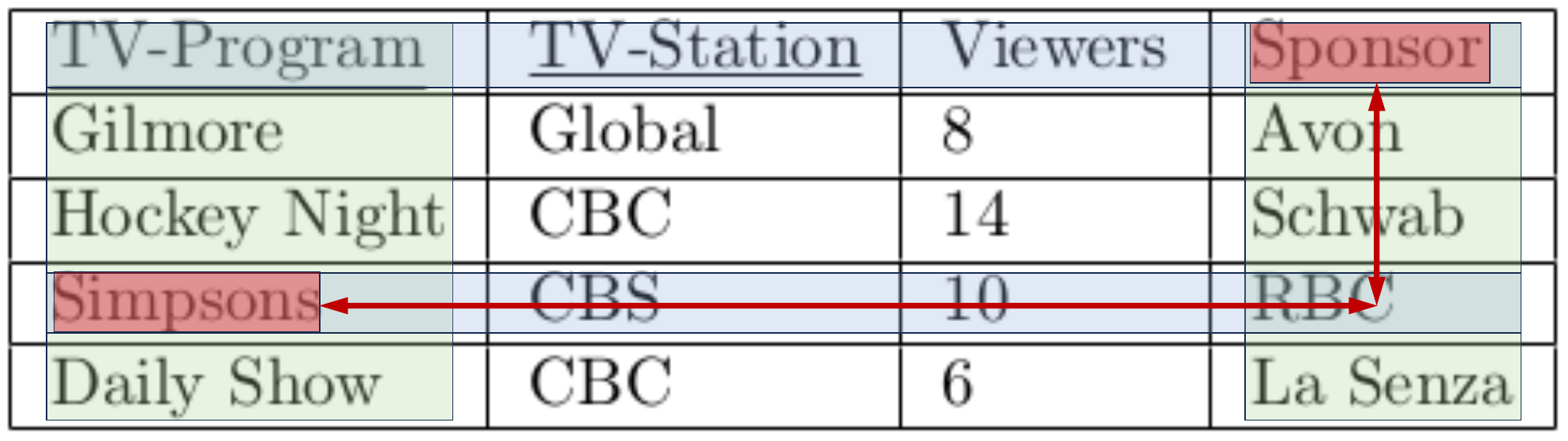}    

\caption{Demonstration of the Logical Distance Prediction task with text region boxes (Red), grid columns (Green), and grid rows (Blue). In this example, the logical distances are both 2 in terms of row and column.}
\label{fig:task}
\end{figure}

\subsection{Logical Distance Prediction Task}
For this task, the model is trained to predict the row and column logical distances between each pair of cells, as illustrated in Figure \ref{fig:task}. In this way, the model learns the ability to understand the basic grids of tables, which serves as the foundation for recognizing complicated table structures. We first pre-process the table images with an off-the-shelf Optical Character Recognition (OCR) system to obtain the 2D position of texts in table cells. Then the horizontal and vertical positions of cells are clustered to conform to the grid rows and columns. In this way, we can obtain the training target of each pair of cells as illustrated in Figure \ref{fig:task}. During the training stage, the features of word region boxes are extracted according to their center points in a similar
fashion as is in Section \ref{sec:reg}. Then the cell features are fed into the logical decoder as introduced in model architecture, where the encoded features are paired for the prediction of logical distance. We employ an L1-loss for the logical distance prediction task.

\section{Experiment}
% TODO: underline to bold figures
% TODO: unifying the name of the metrics

In this section, we conduct comprehensive experiments to research and answer three key questions: 
\begin{itemize}
    \item Is the proposed LORE able to effectively predict the logical locations of table cells from input images? 
    \item Does the LORE framework, modeling TSR as logical location regression, overcome the limitations and cover the abilities of other paradigms?
    \item Is the proposed pre-training strategy beneficial and how does it affect the performance of LORE++.
\end{itemize}

For the first question, we compare LORE with baselines directly predicting logical locations \cite{xue2019res2tim, xue2021tgrnet}. To the best of our knowledge, these are the only two methods that focus on directly predicting the logical locations. Furthermore, we provide a detailed ablation study to validate the effectiveness of the main components. For the second question, we compare LORE with methods that model table structure as cell adjacency or markup sequence with both insights and quantitative results. Finally, we evaluate and analyze the performance of LORE++ to validate the effectiveness of the pre-training method.

\subsection{Datasets} 
\subsubsection{Evaluation Benchmarks}
We evaluate LORE on a wide range of benchmarks, including tables in digital-born documents, i.e., ICDAR-2013 \cite{gobel2013icdar}, SciTSR-comp \cite{chi2019complicated}, PubTabNet \cite{zhong2020image}, TableBank \cite{li-etal-2020-tablebank} and TableGraph-24K \cite{xue2021tgrnet}, as well as tables from scanned documents and photos, i.e.,  ICDAR-2019 \cite{gao2019icdar} and WTW \cite{long2021parsing}. Details of datasets are available in section 2 of the supplementary. It should be noted that ICDAR-2013 provides no training data, so we extend it to the partial version for cross-validation following previous works \cite{raja2020table, liu2021neural, liu2021show}. When training LORE on the PubTabNet, we randomly choose 20,000 images from its training set for efficiency.

\subsubsection{Pre-training Dataset}
For the pre-training of LORE++, we use large-scale table collections such as PubTables1M\cite{smock2022pubtables}, Tablebank, and other small-scale table datasets. The text region, grid rows, and columns are obtained by applying an off-the-shelf OCR system and clustering method. The pre-training set contains 1.5 million table images.

\subsection{Implementation}
LORE is trained and evaluated on table images with the max side scaled to a fixed size of $1024$ ($512$ for SciTSR and PubTabNet) and the short side resized equally. The model is trained for 100 epochs, and the initial learning rate is chosen as $1 \times 10^{-4}$, decaying to $1 \times 10^{-5}$ and $1 \times 10^{-6}$ at the 70th and 90th epochs for all benchmarks. All the experiments are performed on the platform with 4 NVIDIA Tesla V100 GPUs. We use the DLA-34 \cite{yu2018deep} backbone, the output stride $R = 4$, and the number of channels $d = 256$. When implementing on the WTW dataset, a corner point estimation is equipped following \cite{long2021parsing}. The number of attention layers is set to 3 for both the base and the stacking regressors. We run the model 5 times and take the average performance.

LORE++ adopts a 12-layer vision transformer encoder with 12-head self-attention, hidden size of 384, and 1536 intermediate size of feed-forward networks to better fit the MAE pre-training and controls a comparable amount of parameters with prevalent vision backbones in TSR framework such as ResNET-50. The number of attention layers in the logical head is set to 3 as in the vanilla LORE. The images are resized to 224. We pre-train the model using Adam optimizer with a batch size of 196 for steps. We use a weight decay of 0.05, $(\beta_1, \beta_2)=(0.0, 0.95)$, a learning rate of 1.5e-4 and we linearly warm up the learning rate over the first 5\% steps. For downstream fine-tuning, the vision backbone is initialized with the parameters of the pre-trained encoder,  while both the base regressor and stacking regressor are initialized with the parameters of the pre-trained logical decoder. Other fine-tuning configurations are identical to that of the vanilla LORE.

\begin{table*}
  \caption{(Left)Comparison with the TSR methods predicting logical locations. The F-1 score here is the metric for cell detection. Bold denote the best. (Right) Comparison with the TSR methods generating markup sequences.  Bold denote the best.}
    \centering
    \begin{tabular}{lcccccccccccclccc}
    \toprule
       \multicolumn{8}{c}{Logical Location Prediction} &  & & \multicolumn{4}{c}{Markup Prediction} \\
    \cmidrule{1-9}
    \cmidrule{11-14}
    Datasets & \multicolumn{2}{c}{ICDAR-13}  & \multicolumn{2}{c}{ICDAR-19}   & \multicolumn{2}{c}{WTW} & \multicolumn{2}{c}{TG24K} &  & Datasets & \multicolumn{1}{c}{PubTabNet}    & \multicolumn{2}{c}{TableBank}  \\

     metric & F-1 & Acc & F-1 & Acc & F-1 & Acc & F-1 & Acc & &  metric  & TEDS & TEDS & BLEU \\
    \midrule
    ReS2TIM & - & 17.4 & - & 13.8 & - & - & - &  -  & & Image2Text & - & -  & 73.8  \\
    TGRNet    & 66.7 & 27.5 & 82.8 & 26.7 & 64.7 & 24.3 & 92.5 & 84.5 &  & EDD & 89.9 & 86.0  & -  \\
    \midrule
    LORE   &  \textbf{97.2} & \textbf{86.8} &  \textbf{90.6} & \textbf{73.2}  & \textbf{ 96.4} & \textbf{82.9} & \textbf{96.1} & \textbf{87.9} &  & LORE & \textbf{98.1} & \textbf{92.3} & \textbf{91.1} \\
    \bottomrule
    \end{tabular}
  
     \label{tab:logical}

\end{table*}

\subsection{Evaluation Metric}
The TSR models of different paradigms are evaluated using different metrics,  including 1) accuracy of logical locations \cite{xue2019res2tim}, 2) BLEU and TEDS \cite{papineni2002bleu, zhong2020image}, and 3) F-1 score of adjacency relationships between cells \cite{gobel2012methodology, gobel2013icdar}. 

For cell logical location evaluation, a table cell coordinate is represented as $(\text{\emph{start-row}}, \text{\emph{end-row}}, \text{\emph{start-col}}, \text{\emph{end-col}})$ as in the ICDAR competition \cite{gobel2013icdar}. The accuracy, that is the proportion of the cells with four coordinate values correctly predicted, is calculated as the metric of cell location evaluation score. BLEU\cite{papineni2002bleu} is widely adopted in the natural language processing community.  TEDS \cite{zhong2020image} models the markup sequence as a tree (graph) and computes the edit distance between the output structure and label. As for evaluating the adjacency relations, we first convert the table to a list of triplets contains a pair of nodes and their adjacency relation (adjacent/ no adjacent), and make a comparison on relations extracted from output structure and ground truth by precision, recall and F1 score. When evaluating markup generation-based methods, BLEU and TEDS are employed. The accuracy of logical locations, BLEU, and TEDS directly reflect the correctness of the predicted structure, while the adjacency evaluation only measures the quality of intermediate results of the structure.

In our experiments, LORE is evaluated under all three types of metrics, since the logical coordinates are complete for representing table structures and can be converted into adjacency matrices and markup sequences by simple and clarified transformations as introduced in Section. When evaluating TEDS, we use the non-styling text extracted from PDF files following \cite{zheng2021global}. We also report the performance of cell spatial location prediction, using the F-1 score under the IoU threshold of 0.5, following recent works \cite{raja2020table, xue2021tgrnet}. In our experiments, We consider a detected cell to be true positive if its IOU with a ground truth cell bounding box is more than 0.5, following \cite{raja2020table,xue2021tgrnet, raja2022visual}.

\subsection{Results on Benchmarks}
\subsubsection{Result of LORE}
First, we compare LORE with models which directly predict logical locations including Res2TIM \cite{xue2019res2tim} and TGRNet \cite{xue2021tgrnet}. We tune the model provided by \cite{xue2021tgrnet} on the WTW dataset to make a thorough comparison. As shown in Table \ref{tab:logical}, LORE outperforms the previous methods remarkably. The baseline methods can only produce passable results on relatively simple benchmarks of digital-born table images from scientific articles, i.e., TableGraph-24K. TGRNet \cite{xue2021tgrnet} detects cells through segmentation of ruling lines, which would struggle with the spanning cells and deformation of tables. Besides, the graph of cells employed by TGRNet which is constructed according to the Euclidean metric introduces biased prior. LORE achieves better performance benefiting from the flexibility of representing table cells as points, and the cascading regressors which model the intrinsic relation among the logical location of cells. 

Then we evaluate LORE on the markup sequence generation scene against Image2Text \cite{li-etal-2020-tablebank} and EDD \cite{zhong2020image}, with the results also derived from the output logical locations of LORE. Specially, since the TableBank dataset does not provide the spatial locations of cells, we implement LORE trained on SciTSR (1/10 the size of TableBank) for the evaluation on it. The results are shown in Table \ref{tab:logical}. Experiment results indicate that LORE is also more effective even if LORE is trained on much fewer samples. This may be because the logical location prediction paradigm tackles the TSR problem in a direct way to model the 2D structure, rather than the circuitous way where the model needs to learn an additional latent transformation from the structure to the noisy markup sequence.

Thirdly, we compare LORE with models mining the adjacency of cells by relation-based metrics: TabStrNet \cite{raja2020table}, LGPMA \cite{qiao2021lgpma}, TOD \cite{raja2022visual}, FLAGNet \cite{liu2021show} and NCGM \cite{liu2021neural}. The adjacency relation results of LORE are derived from the output logical locations as mentioned before. The results are shown in Table \ref{tab:adjacency}. It is worth noting that LORE performs much better on challenging benchmarks such as ICDAR-2019 and WTW with scanned documents and photos. Tables in these datasets are with more spanning cells and distortions \cite{liu2021neural, long2021parsing}. Experiments demonstrate that LORE is capable of predicting adjacency relations, as by-products of regressing the logical locations.    

\subsubsection{Result of LORE++} Finally, we explore the contrasts of LORE++ and LORE in terms of both spatial location prediction and logical location prediction in Table \ref{tab:lore++}. It indicates that the pre-training triggers the model potential since both tasks are consistently boosted, even if previous models have achieved high performances. Specifically, LORE++ improves the logical location prediction accuracy on ICDAR-2013 the most significantly. This is the smallest dataset with only 158 samples, which illustrates the pre-training stage enhances the generalization of the model. Even though the pre-training dataset contains mostly images of simple digital-born tables, LORE++ is significantly boosted from 82.9\% to 84.1\% on challenging wild dataset of WTW. We utilize the MAE and logical distance prediction task to guide the model to comprehend the general visual clues of tables and logical relationships. As a result, LORE++ demonstrates improvements across diverse datasets.

\begin{table*}[t]
\caption{Comparison with the TSR methods predicting adjacency of cells. The precision, recall, and F-1 score are evaluated on adjacency relationship-based metrics. Bold denote the best.}

\centering
\begin{tabular}{lccccccccccccccc|}
\toprule
Datasets & \multicolumn{3}{c}{ICDAR-13}   & \multicolumn{3}{c}{SciTSR-comp}  & \multicolumn{3}{c}{ICDAR-19}   & \multicolumn{3}{c}{WTW}     \\
 metric & P & R & F1 & P & R & F1 & P & R & F1 & P & R & F1  \\
 
\midrule
\noalign{\smallskip}

TabStrNet & 93.0 & 90.8 & 91.9  & 90.9 & 88.2 & 89.5 & 82.2  & 78.7 & 80.4 & - & - & -  \\
LGPMA    & 96.7 & 99.1 & 97.9  & 97.3 & 98.7 & 98.0 & - & - & - & - & - & - \\
TOD   & 98.0 & 97.0 & 98.0  & 97.0 & 99.0 & 98.0 & 77.0 & 76.0 & 77.0 & - & - & -  \\
FLAGNet & 97.9 & \textbf{99.3} & 98.6  & 98.4 & 98.6 & 98.5 & 85.2 & 83.8 & 84.5 & 91.6  & 89.5  & 90.5   \\
NCGM & 98.4 & \textbf{99.3} & 98.8  & 98.7 & 98.9 & 98.8 & 84.6 & 86.1 & 85.3 & 93.7 & 94.6 & 94.1  \\
\midrule
LORE & \textbf{99.2} & 98.6 & \textbf{98.9} & \textbf{99.4} & \textbf{99.2} & \textbf{99.3} & \textbf{87.9} & \textbf{88.7} & \textbf{88.3} & \textbf{94.5} & \textbf{95.9} & \textbf{95.1} \\
\bottomrule
\end{tabular}

\label{tab:adjacency}

\end{table*}

\begin{table*}[t]
\caption{Comparison with the TSR methods predicting adjacency of cells. The precision, recall and F-1 score are evaluated on adjacency relationship-based metrics. Bold denote the best.}

\centering
\begin{tabular}{lccccccccccccccc|}
\toprule
Datasets & \multicolumn{3}{c}{ICDAR-13}   & \multicolumn{3}{c}{SciTSR-comp}  & \multicolumn{3}{c}{PubTabNet}   & \multicolumn{3}{c}{WTW}     \\
 metric & D-F1 & R-F1 & Acc & D-F1 & R-F1 & Acc & D-F1 & R-F1 & Acc & D-F1 & R-F1 & Acc  \\
 
\midrule
\noalign{\smallskip}

FLAGNet & - & 98.6 & -  & - & 98.5 & - & - & - & - & -  & 90.5  & -   \\
NCGM & - & 98.8 & -  & - & 98.8 & - & - & - & - & - & 94.1 & -  \\
\midrule
LORE &   97.2  & 98.9 & 86.8 & 
97.1 & 99.3 & 94.6 & 
92.4 & 98.7 & 91.0 & 
96.4 & 95.1 & 82.9 \\
LORE++ &   \textbf{98.5}  & \textbf{99.2} & \textbf{93.2} & 
\textbf{99.1} & \textbf{99.4} & \textbf{95.7} & 
\textbf{94.4} & \textbf{99.1} & \textbf{92.7} & 
\textbf{97.0} & \textbf{96.9} & \textbf{84.1} \\
\bottomrule
\end{tabular}
\label{tab:lore++}
\end{table*}

\begin{table*}
    \caption{Ablation study of LORE. A-c, A-r, and Acc refer to the accuracy of column indices, row indices, and all logical indices. All these models are trained from scratch according to the `Implementation' section.}
    \centering
    \setlength\tabcolsep{0.27cm}
    \begin{tabular}{lcccccccccc}
    \toprule
    
 \multirow{2}{*}{N}   & \multicolumn{3}{c}{Objectives}  & \multirow{2}{*}{Cascade} & \multicolumn{3}{c}{Architecture}  &  \multicolumn{3}{c}{Metrics}    \\

     & $L_1$ & Inter & Intra &  &Encoder & Base & Stacking  & A-c &  A-r & Acc  \\
    %  & $L_1$ & Inter & Intra &  &Encoder & Base & Stacking  & ${\rm A}_c$ &  ${\rm A}_r$  & Acc  \\
    
  \midrule
  1a & \checkmark & - & - & \checkmark & Attention &3 & 3 & 87.2 &  84.8 & 79.4  \\
  1b & \checkmark & \checkmark & - & \checkmark & Attention &3 & 3 & 87.6 &  86.6 & 80.2  \\
  1c &\checkmark & - & \checkmark & \checkmark & Attention &3& 3  & 89.5 & 87.1 & 81.2  \\
  1d & \checkmark & \checkmark & \checkmark & \checkmark & Attention &3 & 3  & 91.3 & 87.9 & 82.9 \\
  \midrule
   
  2a & \checkmark & \checkmark & \checkmark & \checkmark &  GNN & 3& 3 & 88.2 & 82.6 & 77.0 \\
  2b & \checkmark & \checkmark & \checkmark & - & Attention & 6 & 0  & 88.7  & 85.3 & 79.8 \\

    \bottomrule
    \end{tabular}

     \label{tab:ablation}
\end{table*}

\subsection{Ablation}
\subsubsection{Ablation Study of LORE} To investigate how the key components of our proposed LORE contribute to the logical location regression, we conduct an intensive ablation study on the WTW dataset. Results are presented in Table \ref{tab:ablation}. First, we evaluate the effectiveness of the inter-cell loss $L_{inter}$ and the intra-cell loss $L_{intra}$, by training several models turning them on and off. According to the results in experiments 1a and 1b, we see that the inter-cell supervision improves the performance by +0.8\%Acc. And from 1a and 1c, the intra-cell supervision benefits more by +1.8\%Acc, for the reason that it makes up the message-passing and aggregating mechanism, which pays less attention to intra-cell relations than inter-cell relations according to its inter-cell nature. The combination of the two supervisions makes the best performance.

% {\fontsize{9pt}{0pt} \selectfont 
Then we evaluate the influence of model architecture, i.e., the pattern of message aggregation and the importance of the cascade framework. In experiment 2a, we replace the self-attention encoder with a graph-attention encoder similar to graph-based TSR models \cite{qasim2019rethinking, xue2021tgrnet} with an equal amount of parameters with LORE. It causes a drop in performance consistently. The graph-based encoder only aggregates information from the top-K nearest features of each node based on Euclidean distance, which is biased for table structure. In Experiment 2b, we use a single regressor of 6 layers instead of two cascading regressors of 3 layers. We can observe a performance degradation of 3.1\%Acc from 1d to 2b, showing that the cascade framework can better model the dependencies and constraints between logical locations of different cells. 

\subsubsection{Ablation Study of Pre-training} The ablation of the pre-training task is in Table \ref{tab:pre-ablation}. The 3a is the baseline of LORE. Actually, replacing the CNN backbone of LORE (3a) with ViT (3c) leads to a performance drop (from 82.9\% to 82.1\%). Perhaps it's because the amount of data for WTW is not sufficient, which leads to the inferior performance of ViT compared to CNN. Here the ViT architecture is employed to cater the MAE pre-training for convenience. The difference between experiment \textless 3a, 3c\textgreater  \, and \textless 3a, 3d\textgreater \, indicates that using only the table data set for MAE pre-training is beneficial, even if the ImageNet is much larger than our pre-training dataset for the pre-trained model learns the basic visual clues of tables, such as cell regions and ruling lines. Adding the logical distance prediction task results in a substantial improvement in logical location prediction according to experiments 3d and 3e. Notably, the spatial prediction task is also boosted by the logical distance prediction task. 

% In addition, it shows relatively lower performance with the original LORE to only pretraining the model with ViT backbone on the ImageNet

\begin{table}
    \caption{Ablation study of LORE++. A-c, A-r, and Acc refer to the accuracy of column indices, row indices, and all logical indices. All these models are trained from scratch according to the `Implementation' section.}
    \centering
    \setlength\tabcolsep{0.13cm}
    \begin{tabular}{cccccccc}
    \toprule
    
 \multirow{2}{*}{N}   & \multicolumn{2}{c}{Task} & \multirow{2}{*}{Backbone} & \multirow{2}{*}{Data}  & \multicolumn{3}{c}{Metrics}     \\

      & MAE & LDP & & & D-F1 & R-F1 & Acc  \\

  \midrule
  3a({\fontsize{5}{11}\selectfont LORE}) & - & - & CNN & ImageNet & 96.4 &  95.1 & 82.9 \\
  \midrule
  3b & - & - & ViT & None & 87.6 &  82.3 & 75.3 \\
  3c & \checkmark & - & ViT & ImageNet & 96.4 &  95.3 & 82.1  \\
  3d & \checkmark & -  & ViT & Ours & 96.9 & 96.4 & 83.2  \\
  3e({\fontsize{5}{11}\selectfont LORE++}) &\checkmark & \checkmark & ViT & Ours & 97.0 & 96.9 & 84.1  \\
  
    \bottomrule
    \end{tabular}

     \label{tab:pre-ablation}
\end{table}

\begin{figure}[!t]
\centering
\captionsetup[subfloat]{font=scriptsize}
\subfloat[Original structure]{\includegraphics[width=0.96\linewidth]{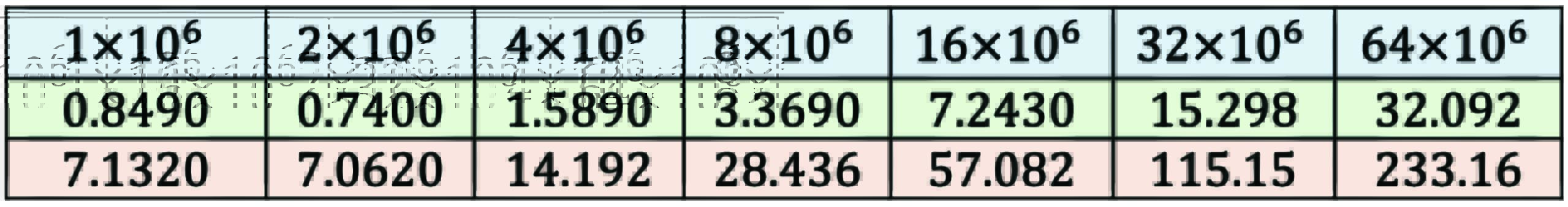}%
\label{fig_first_case}}
\hfil
\subfloat[Shifted structure]{\includegraphics[width=0.96\linewidth]{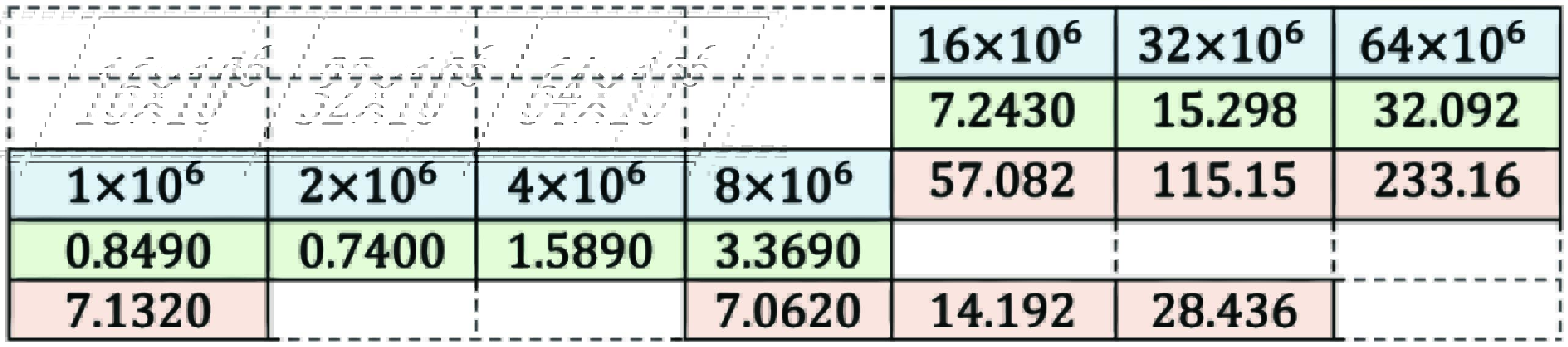}%
\label{fig_second_case}}
\caption{An example of a severely shifted structure. Its adjacency-relationship F-1 is 84\%, while the logical location accuracy is just 43\%.}
\label{fig:shift}
\end{figure}

\subsection{Further Comparison among Paradigms}
In this section, we further compare models of different TSR paradigms introduced before. Previous methods that predict logical locations lack a comprehensive comparison and analysis between these paradigms. We demonstrate how LORE overcomes the limitations of the adjacency-based and the markup-based methods by controlled experiments.

The adjacency of cells alone is not sufficient to represent table structures. Previous methods employ heuristic rules based on spatial locations \cite{liu2021neural} or graph optimizations \cite{qasim2019rethinking} to reconstruct the tables. However, it takes tedious modification to make the pre-defined parts compatible with datasets of different types of tables and annotations. Furthermore, the adjacency-based metrics sometimes fail to reflect the correctness of table structures, as depicted in Figure \ref{fig:shift}. Experiments are conducted to verify this argument quantitatively. We turn the linear layer of the stacking regressor of LORE into an adjacency classification layer of paired cell features and employ post-processings as in NCGM \cite{liu2021neural} to reconstruct the table. The results are in Table \ref{tab:rflogi}. Although this modified model (Adj. paradigm) achieves competitive results with state-of-the-art baselines evaluated on adjacency-based metrics, the accuracy of logical locations obtained from heuristic rules decreases obviously compared to LORE (Log. paradigm), especially on WTW, which contains more spanning cells and distortions.

\begin{figure}[!t]
\centering
\captionsetup[subfloat]{font=scriptsize}
\subfloat[Attention activation of the base regressor]{\includegraphics[width=0.96\linewidth]{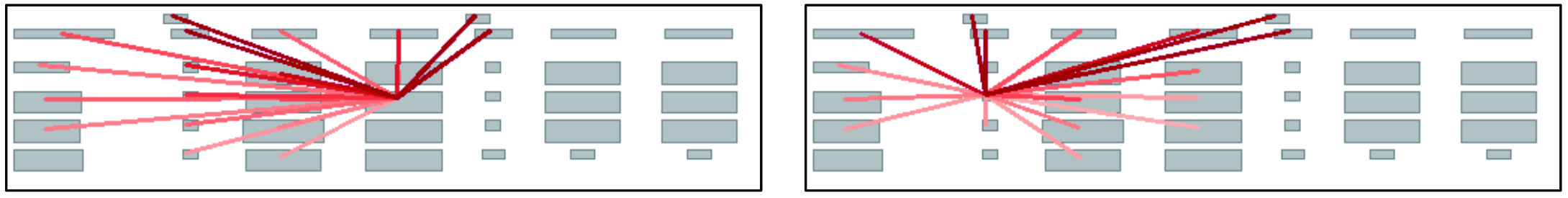}%
\label{fig_first_case}}
\hfil
\subfloat[Attention activation of the stacking regressor]{\includegraphics[width=0.96\linewidth]{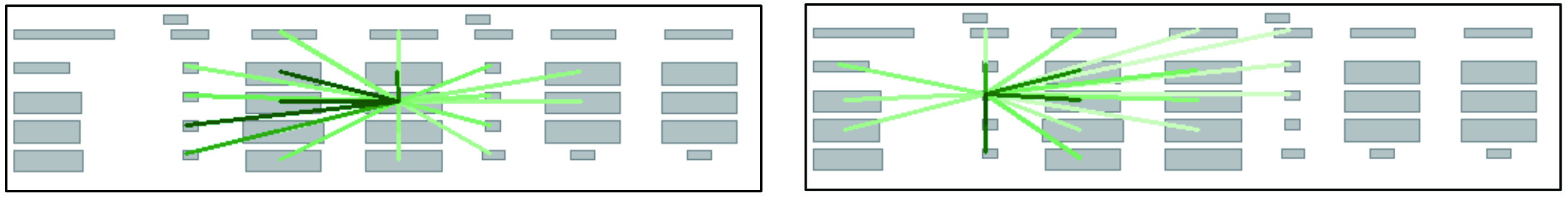}%
\label{fig_second_case}}
\hfil
\subfloat[Attention activation of the non-cascade regressor]{\includegraphics[width=0.96\linewidth]{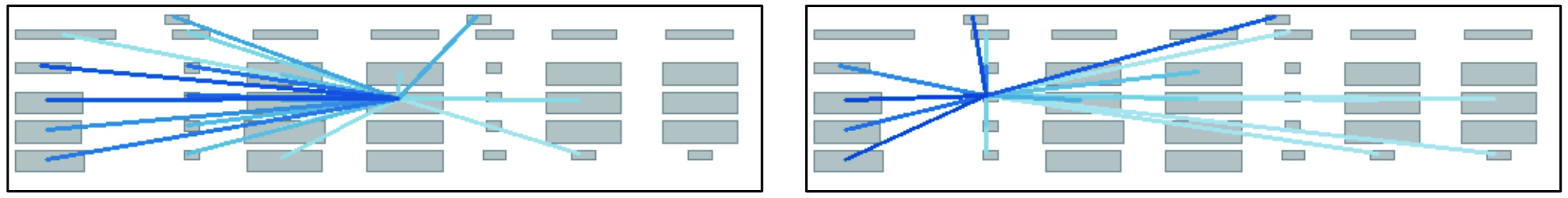}%
\label{fig_second_case}}
\caption{Visualization of the self-attention weights in the cascade and non-cascade regressors for two table cells. Text masks represent table cells and only top-20 weights are visualized for clarity.}
\label{fig:att}
\end{figure}

\begin{table}
 \caption{Evaluation results of the adjacency and the logical location paradigms. A-all and A-sp refer to the logical location accuracy of all cells and spanning cells (more than one row/column). Sci-c denotes SciTSR-comp. }
    \centering
    \setlength \tabcolsep{0.19cm}
    \begin{tabular}{lccccccc}
    \toprule
    \multirow{2}{*}{Data} & \multirow{2}{*}{Paradigm} & \multicolumn{3}{c}{Adj. Metrics}    & \multicolumn{3}{c}{Log. Metrics} \\

    % & & P & R & F-1 &   {${\rm Acc}_{all}$}   & {${\rm Acc}_{sp}$} \\
    & & P & R & F-1 &   A-all   & A-sp \\
    \midrule
    \multirow{2}{*}{Sci-c} & Adj. & 98.6 & 98.9 & 98.7 & 94.7 & 63.5\\
    & Log. & 99.4 & 99.2  &  99.3  & 97.3 & 87.7 \\
    \midrule
    \multirow{2}{*}{WTW} &Adj. & 95.0 & 93.7  &  94.3  & 51.9 & 20.2 \\
    & Log. & 94.5 & 95.9  &  95.1  & 82.9 & 63.8\\
  
    \bottomrule
    \end{tabular}
   
     \label{tab:rflogi}
     %It can be observed that the heuristic rules struggle with complex table structures.

\end{table}

The markup-sequence-based models leverage image encoders and sequence decoders to predict the label sequences. Since the markup language has plenty of control sequences formatting styles, they can be viewed as noise in labels and impede model training \cite{xue2021tgrnet}. It requires much more training samples and computational costs. As shown in Table \ref{tab:consum}, the number of training samples of the EDD model on the PubTabNet dataset is more than ten times larger than that of both LORE and LORE++. Besides, the inference process is rather time-consuming (See Table \ref{tab:consum}) due to the sequential decoding pattern, while models of other paradigms compute for each cell in parallel. The average inference time is computed from the validation set of PubTabNet with the images resized to $1280\times1280$ for both models.

\subsection{Further Analysis on Cascade Regressors}
We conduct experiments to investigate the effect of the cascade framework on the prediction of logical coordinates. In Figure \ref{fig:att}, we visualize the attention maps of the last encoder layer of the cascade/single regressors of two cells, i.e., the models 1d and 2b in Table \ref{tab:ablation}. In the cascade framework, the base regressor in Figure \ref{fig:att} (a) focuses on the heading cells (upper or left) to compute logical locations. While the stacking regressor in Figure \ref{fig:att} (b) pays more attention to the surrounding cells to discover finer dependencies among logical locations and make sure the prediction is subject to natural constraints, which is in line with human intuition when designing a table. However, the non-cascade regressor in Figure \ref{fig:att} (c) can only play a role similar to the base regressor, which leaves out important information for the prediction of logical locations.

% \begin{table}
% \caption{Comparison of LORE and the markup generation model EDD in terms of training samples and average inference time.}
%     \centering
%     \setlength\tabcolsep{0.19cm}
%     \begin{tabular}{lcc}
%     \toprule
%      & \#Train Samples    & Inference Time  \\
%     \midrule
%     EDD & 339000 & 14.8s\\
%     LORE & 20000 & 0.45s \\
  
%     \bottomrule
%     \end{tabular}
    
%      \label{tab:consum}

% \end{table}

\begin{table}
\caption{Comparison of LORE and the markup generation model EDD in terms of training samples and average inference time.}
    \centering
    \setlength\tabcolsep{0.25cm}
    \begin{tabular}{lccc}
    \toprule
      & EDD & LORE & LORE++  \\
    \midrule
    \#Train Samples & 339K & 20K & 20K \\
    Inference Time & 14.8s & 0.45s & 0.43s \\
  
    \bottomrule
    \end{tabular}
    
     \label{tab:consum}

\end{table}

\begin{table}
 \caption{Computational Analysis. The units are million for the number of parameters and giga for the FLOPs. }
    \centering
    \begin{tabular}{lcccc}
    \toprule
     & DLA-34 & LORE & LORE++ \\
    \midrule
    \#Params & 15.9  & 24.2 & 29.7\\
    FLOPs & 74.6  & 75.2 & 88.3 \\
  
    \bottomrule
    \end{tabular}
   
     \label{tab:flops}

\end{table}

\begin{table}[t]
 \caption{Evaluation results of the adjacency and the logical location paradigms. A-all and A-sp refer to the logical location accuracy of all cells and spanning cells (more than one row/column). Sci-c denotes SciTSR-comp. }
    \centering
    \begin{tabular}{lcccccc}
    \toprule
    \multirow{2}{*}{Model}  & \multicolumn{3}{c}{SciTSR-COMP}    & \multicolumn{3}{c}{ICDAR2013} \\

    % & & P & R & F-1 &   {${\rm Acc}_{all}$}   & {${\rm Acc}_{sp}$} \\
     & D-F1 & R-F1 & Acc & D-F1 & R-F1 & Acc \\
    \midrule
    LORE & 92.9 & 96.4 & 87.1 & 92.1 & 93.2 & 78.6 \\
    LORE++ & \textbf{95.4} & \textbf{97.9} & \textbf{93.5} & \textbf{95.5} & \textbf{98.6} & \textbf{87.5} \\
    
    \bottomrule
    \end{tabular}
   
     \label{tab:gen}
     %It can be observed that the heuristic rules struggle with complex table structures.

\end{table}

\subsection{Computational Analysis}
We summarize the model size and the inference operations of LORE and LORE++ in Table \ref{tab:flops}, with the input images at $1024\times1024$ and the number of cells as 32.  It is observed that the complexity of LORE is at an equal level to a key point-based detector \cite{zhou2019objects} with the same backbone, showing the efficiency of LORE. The LORE++ is relatively larger since the ViT backbone is employed, but the model size maintains a similar level to the original LORE. Besides, the ablation in Table \ref{tab:pre-ablation} has validated that the improvements are not owing to the different backbone networks.

\begin{figure*}[t]
\vspace{-2mm}
  \centering
\includegraphics[width=2\columnwidth]{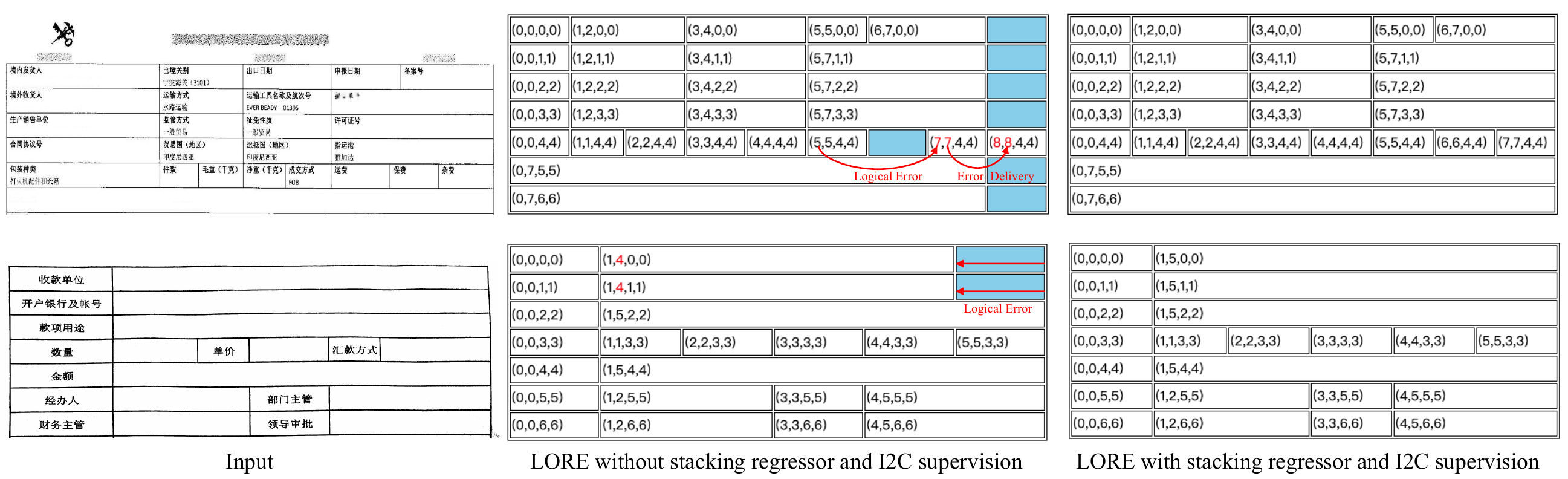}    
\vspace{-2mm}
\caption{Comparison of the TSR output without/with stacking regressor and I2C supervisions. The 4-tuple in each cell represents the logical location of the cell, i.e. the starting-column, ending-column, starting-row, and ending-row. Colorized cells are empty cells that occur when converting model results to a spreadsheet due to erroneous logical coordinates during model inference. The red coordinates indicate incorrect inference results.}
\label{fig:visual_case}
\vspace{-2mm}
\end{figure*}

\subsection{Data Efficiency}

\begin{figure}[t]
\vspace{-2mm}
  \centering
\includegraphics[width=0.98\columnwidth]{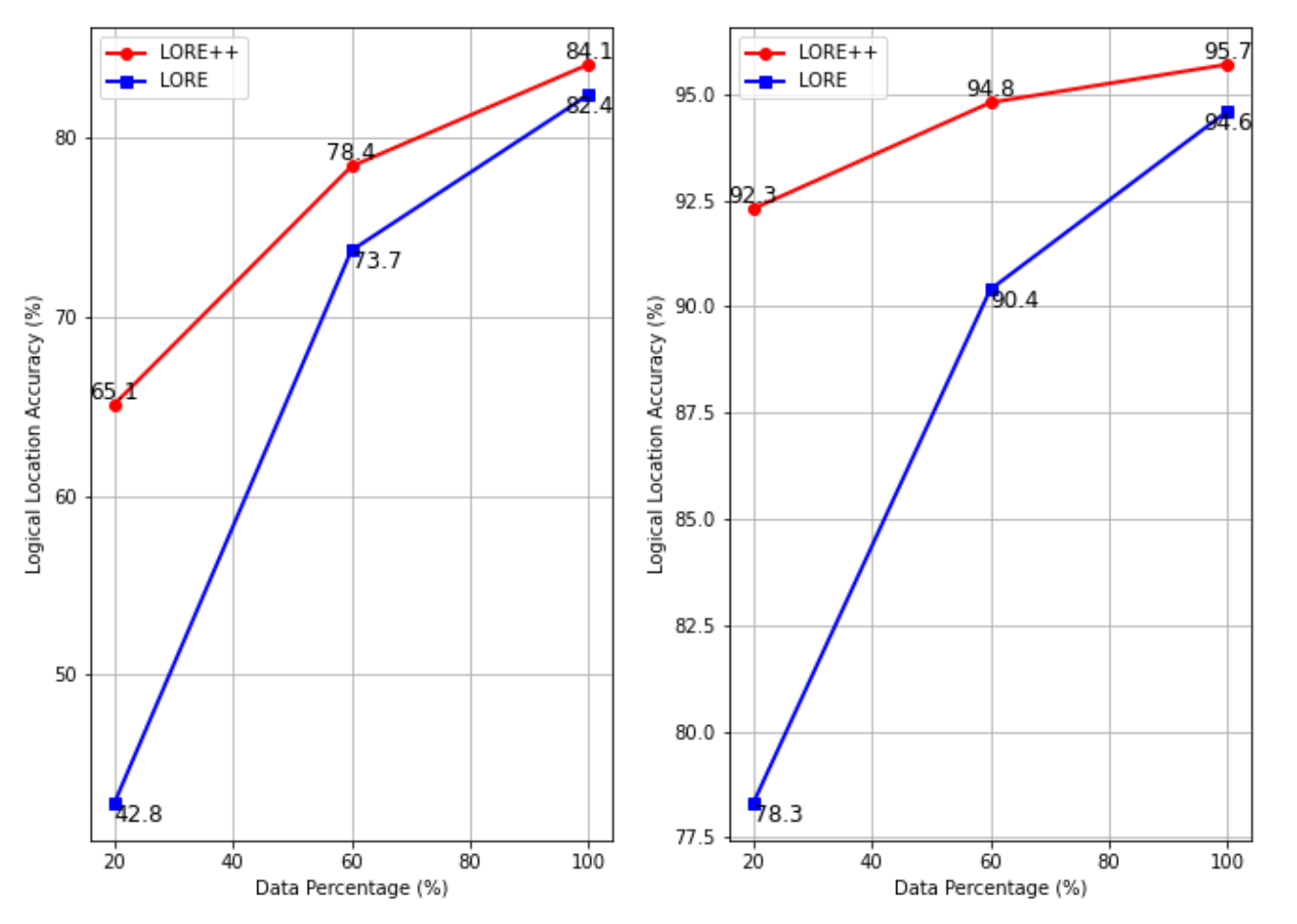}   
\vspace{-2mm}
\caption{Comparison of data efficiency of LORE++ and LORE on WTW and SciTSR-comp dataset.  }
\label{fig:de}
\vspace{-3mm}
\end{figure}

To validate the effectiveness of pre-training LORE in improving data efficiency, we compare the pre-trained LORE++ model with the baseline model LORE at different training settings: training them using 20\%, 60\%, and 100\% training sets of WTW and SciTSR for equivalent 100 epochs regarding the full training set, e.g., we employ 5 times epochs when using 20\% data for training compared with using 100\% data. The results are shown in Figure \ref{fig:de}. As can be seen, LORE++ consistently outperforms the LORE baseline by a large margin in terms of data efficiency. LORE++ using 60\% training data achieves comparable performance with LORE using all data on the SciTSR dataset.  With the proposed proxy tasks, the pre-trained LORE++ has a grasp of the notion of basic vision and logical clues, which makes learning the TSR task more efficient with less training data. 

\subsection{Generalization}
In this section, we conduct experiments to validate whether the pre-training enhances the generalization ability of LORE. We train the model on a hybrid dataset which contains the WTW training set, 20,000 samples of the PubTabNet training set, and the TableGraph24K training set, and evaluate this model on the ICDAR2013 and SciTSR-comp datasets. The results are displayed in Table \ref{tab:gen}, which depicts the generalization ability is obviously boosted after pretraining.

\subsection{Visualization of TSR Results}
In order to reveal the effectiveness of considering the interaction among logical locations, we visualize the structure recognition results of LORE models without/with a stacking logical regressor and the I2C losses. As depicted in Figure \ref{fig:visual_case}, models without the stacking regressor and I2C losses encounters problem when predicting the logical location of complicated structures and blank cells, such as the logical errors marked as red lines in Figure \ref{fig:visual_case}. While the model with a stacking logical regressor and the I2C losses fixes these errors owning to the stacking regressor refining a rough results of logical locations and knowledge learned from the constrains among logical location of cells.

\section{Conclusion}

In summary, we propose LORE, a TSR framework that effectively regresses the spatial locations and the logical locations of table cells from the input images. Furthermore, it models the dependencies and constraints between logical locations by employing the cascading regressors along with the inter-cell and intra-cell supervisions. LORE is straightforward to implement and achieves competitive results, without tedious post-processing or sequential decoding strategies. Experiments show that LORE outperforms state-of-the-art TSR methods under various metrics and overcomes the limitations of previous TSR paradigms. Additionally, we propose the pre-training method of LORE, resulting in an upgraded version called LORE++, which outperforms the baseline LORE in terms of accuracy and data efficiency.

\section{Acknowledgement}
The authors would like to thank the Editors and Reviewers for their hard work and valuable comments.
%\newpage

\bibliographystyle{IEEEtran}
\bibliography{reference}

\vfill

\end{document}